\documentclass[runningheads]{llncs}
\pdfoutput=1
\usepackage{hyperref}
\hypersetup{
  pdfinfo={
    Title={Deep Learning for Semantic Segmentation on Minimal Hardware},
    Author={Sander G. van Dijk, Marcus M. Scheunemann},
    Subject={This paper presents an approach based on semantic segmentation for processing full VGA images in real-time on a low-power mobile processor. It can further handle multiple image dimensions without retraining, it does not require specific domain knowledge for achieving a high frame rate and it is applicable on a minimal mobile hardware.},
    Keywords={deep learning, semantic segmentation, mobile robotics, computer vision, minimal hardware},
    Creator={Some PDFLatex installation of arXiv},
    Lang = {en-GB}
  }
}

\usepackage{graphicx}
\usepackage[dvipsnames]{xcolor}
\usepackage{tikz}
\usepackage{pgfplots}
\usepackage{pgf-umlsd}
\usepackage{ifthen}
\usepackage{booktabs}

\usepackage[range-units=single, list-units=single]{siunitx}
	\DeclareSIUnit\fps{\text{fps}}
	\DeclareSIUnit\fpstext{\text{frames per second}}

\begin{document}

\title{
Deep Learning for Semantic Segmentation\\ on Minimal Hardware}

\author{Sander G. van Dijk \and Marcus M. Scheunemann}

\institute{University of Hertfordshire, AL10 9AB, UK}

\maketitle

\begin{abstract}
  Deep learning has revolutionised many fields, but it is still
  challenging to transfer its success to small mobile robots with
  minimal hardware. Specifically, some work has been done to this
  effect in the RoboCup humanoid football domain, but results that are
  performant and efficient and still generally applicable outside of 
  this domain are lacking. We propose an approach conceptually different 
  from those taken previously. It is based on semantic segmentation and 
  does achieve these desired properties. In detail, it is being able to process full VGA 
  images in real-time on a low-power mobile processor. It can further handle multiple image 
  dimensions without retraining, it does not require specific domain knowledge for achieving a high 
  frame rate and it is applicable on a minimal mobile hardware.
\end{abstract}

\section{Introduction}
\label{sec:introduction}

Deep learning~(DL) has greatly accelerated progress in many areas of
artificial intelligence~(AI) and machine learning. Several breakthrough ideas and methods,
combined with the availability of large amounts of data and
computation power, have lifted classical artificial neural networks~(ANNs) 
to new heights in natural language processing, time series
modelling and advanced computer vision problems~\cite{lecun2015deep}. 
For computer vision in particular, networks using convolution operations, 
i.e., \emph{Convolutional Neural Networks}~(CNNs), have had great success.

Many of these successful applications of DL rely on cutting edge
computation hardware, specifically high-end GPU processors, sometimes
in clusters of dozens to hundreds of
machines~\cite{silver2016mastering}. Low-power robots, such as the
robotic footballers participating in RoboCup, are not able to carry
such hardware. It is not a surprise that the uptake of DL in the domain
of humanoid robotic football has lagged behind. Some
demonstrations of its use became
available recently~\cite{speck2016ball,albani2016deep,cruz2017using,hess2017stochastic,schnekenburger2017detection}.
However, as we will discuss in the next section, these applications are so far rather
limited; either in terms of performance or in terms of their generalisability
for areas other than RoboCup.

In this paper, we will address these issues and present a DL framework
that achieves high accuracy, is more generally applicable and still
runs at a usable frame rate on minimal hardware.

The necessary conceptual switch and main driver behind these results
is to apply DL to the direct semantic segmentation of camera images,
in contrast to most previous work in the humanoid robotic football
domain that has applied it to the final object detection or
recognition problem. Semantic segmentation is the task of assigning a
class label to each separate pixel in an image, in contrast to
predicting a single output for an image as a whole, or some sub-region
of interest. There are three primary reasons why this approach is
attractive.

Firstly, semantic segmentation networks can be significantly smaller
in terms of learnable weights than other network types. The number of
weights in a convolution layer is reduced significantly compared to
the fully connected layers of classical ANNs, by `reusing' filter
weights as they slide over an image. However, most image
classification or object detection networks still need to convert a 2D
representation into a single output, for which they do use fully
connected layers on top of the efficient convolution layers. The
number of weights of fully connected layers is quadratic in their
size, which means they can be responsible for a major part of the
computational complexity of the network. Semantic segmentation
networks on the other hand typically only have convolution
layers---they are \emph{Fully Convolutional Networks}~(FCNs)---and so
do away with fully connected ones, and the number of their weights
only grows linearly with the number of layers used.

Secondly, the fully convolutional nature also ensures that the network is
independent of image resolution. The input resolution of a network
with a fully connected output layer is fixed by the number of weights
in that layer. Such a network, trained on data of those dimensions,
cannot readily be reused on data of differing sizes; the user will
have to crop or rescale the input data, or retrain new fully connected
layers of the appropriate size. Convolution operations on the other
hand are agnostic of input dimensions, so a fully convolutional
network can be used at any input resolution\footnote{Given that the 
input dimensions are not so small that any down-sampling operations, 
e.g. max pooling, would reduce the resolution to nil.}. This
provides very useful opportunities. For example, if a known
object is tracked, or an object is known to be close to the camera, 
the algorithm allows for an on-demand up and down scaling of vision effort. 
Instead of processing a complete camera frame when searching 
for such an aforementioned object, only an image subset or a downscaled 
version of a camera frame is processed.

Finally, semantic segmentation fits in directly with many popular
vision pipelines used currently in the RoboCup football
domain. Historically, the domain consisted of clearly colour coded
concepts: green field, orange ball, yellow/blue goalposts. Commonly a
lookup-table based approach is used to label each pixel separately,
after which fast specialised connected component, scanning, or
integral image methods are applied to detect and localise all relevant
objects. Over the years the scenario has developed to be more
challenging (e.g., natural light, limited colours) and unstructured, 
making the lookup-table methods less feasible. Using a semantic segmentation 
CNN that can learn to use more complex features would allow the simple 
replacement of these methods and still reuse all highly optimised 
algorithms of the existing vision pipeline.

\section{Related Work}
\label{sec:related_work}

The RoboCup domain offers a long history in research on efficient,
practical computer vision methods; already the very first RoboCup
symposium in 1997 dealt with ``Real-Time Vision Processing for a
Soccer Playing Mobile Robot``~\cite{cheng1997real}, and the 2016
Kid-Size champions still heavily relied on optimisations, e.g.,
regions of interest and downscaled images, to make vision
viable~\cite{allali2017rhoban}. To ensure keeping track of 
and participating in a dynamic game of football,
the robots ideally should process at least \SIrange{20}{30}{\fps}. However, they only
have very limited energy resources yielding minimal computational
power for achieving that.

Recent developments in low-power, mobile computational platforms, as well
as in efficient deep learning, have now made it possible to adopt DL in
small mobile robots. One of the first works on DL in the RoboCup
domain presented a CNN trained to separately predict the $x$ and $y$
coordinates of a ball in a camera image~\cite{speck2016ball}. Although
this network performed relatively well, it could only process a few
images per second and operated on heavily downscaled images. At the
same time other authors were able to create a CNN-based system that
could recognise Nao robots within milliseconds~\cite{albani2016deep}.
However, this method relied on a region proposal preprocessing method
very specific to RoboCup. This work was later generalised to a
different RoboCup league~\cite{javadi2017humanoid}, but still
relies on the specifically colour-coded nature of the robot football
pitch. Instead, the approach taken in this paper is to use CNNs as the
very first processing step, and only \emph{after} that step apply
domain specific algorithms. This same approach was taken in a recent
work very much related to ours~\cite{schnekenburger2017detection}, but
for large humanoid robots with powerful hardware that cannot feasibly
be used by smaller size mobile robots, such as Kid-Size humanoids or
perhaps drones.

In recent years there has been a growing body of work on creating
small, but capable networks, for enabling their use on more restricted
hardware. One approach is to try to minimise the complexity of
convolutions by first applying a `bottleneck' \num{1x1}~convolution layer
that reduces the number of features used in the actual $\mathrm{N}\times \mathrm{N}$~convolution. 
This idea originated with ResNet~\cite{he2016deep} to help
make training of very deep networks feasible, but at the same time can
also reduce run time costs. For networks of the sizes used in this
paper however, the computational cost of a bottleneck layer outweighed
the benefit of a reduced number of features in the subsequent layer. A
different idea is to discretise quantities used in the networks, with
the idea that integer operations can be much more efficient in low-end
computation devices. The culmination of this idea is in network
designs such as XNOR-nets~\cite{rastegari2016xnor} that use very basic
and fast bitwise operations during prediction, and
DoReFa-nets~\cite{zhou2016dorefa} that further extend this idea to
training. We do not study such binary nets here, as at the moment
there is no implementation available of the operators required by such
networks for the most popular deep learning libraries, and we are
interested in systems that can be easily adapted and implemented by
the reader using such libraries.

Instead, the optimisations applied to our networks are very much
motivated by MobileNets~\cite{howard2017mobilenets}. Most notably, we
utilise \emph{depthwise separable convolutions} to significantly
reduce the computational complexity of our segmentation networks. Such
convolutions split a regular convolution into a filter and a
combination step: first a separate 2D filter is applied to each input
channel, after which a \num{1x1}~convolution is applied to combine the
results of these features. This can be seen as a factorisation of a
full convolution that reduces the computational cost by a factor of
$\frac{1}{N}+\frac{1}{K^2}$, where $N$ is the number of output
features and $K$ the kernel size. Not all convolutions can be
factorised like this, so separable convolutions have less expressive
power, but the results of the original MobileNets and those reported
here show they can still perform at high accuracy.

\section{Network Architecture}
\label{sec:network_architecture}

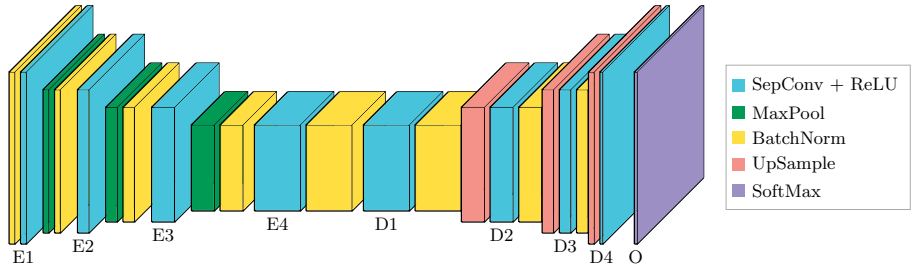
\begin{figure}[t]
  \noindent\resizebox{\textwidth}{!}{
    \begin{tikzpicture}[baseline = (current bounding box.west)]

      \newcommand{\networkLayer}[6]{
        \def\a{#1} %
        \def\b{0.02}
        \def\c{#2} %
        \def\t{#3} %
        \ifthenelse {\equal{#6} {}} {\def\y{0}} {\def\y{#6}} %

        \draw[line width=0.25mm](\c+\t,0,0) -- (\c+\t,\a,0) -- (\t,\a,0);                                                      %
        \draw[line width=0.25mm](\t,0,\a) -- (\c+\t,0,\a) node[midway,below] {#5} -- (\c+\t,\a,\a) -- (\t,\a,\a) -- (\t,0,\a); %
        \draw[line width=0.25mm](\c+\t,0,0) -- (\c+\t,0,\a);
        \draw[line width=0.25mm](\c+\t,\a,0) -- (\c+\t,\a,\a);
        \draw[line width=0.25mm](\t,\a,0) -- (\t,\a,\a);

        \filldraw[#4] (\t+\b,\b,\a) -- (\c+\t-\b,\b,\a) -- (\c+\t-\b,\a-\b,\a) -- (\t+\b,\a-\b,\a) -- (\t+\b,\b,\a); %
        \filldraw[#4] (\t+\b,\a,\a-\b) -- (\c+\t-\b,\a,\a-\b) -- (\c+\t-\b,\a,\b) -- (\t+\b,\a,\b);

        \ifthenelse {\equal{#4} {}}
        {} %
        {\filldraw[#4] (\c+\t,\b,\a-\b) -- (\c+\t,\b,\b) -- (\c+\t,\a-\b,\b) -- (\c+\t,\a-\b,\a-\b);} %
      }

      \networkLayer{3.0}{0.1}{-0.2}{color=Goldenrod}{}{}
      \networkLayer{3.0}{0.1}{0.0}{color=SkyBlue}{E1}{}
      \networkLayer{2.5}{0.1}{0.2}{color=ForestGreen}{}{}
      \networkLayer{2.5}{0.1}{0.4}{color=Goldenrod}{}{}
      
      \networkLayer{2.5}{0.2}{0.8}{color=SkyBlue}{E2}{}
      \networkLayer{2.0}{0.2}{1.1}{color=ForestGreen}{}{}
      \networkLayer{2.0}{0.2}{1.4}{color=Goldenrod}{}{}
      
      \networkLayer{2.0}{0.4}{1.9}{color=SkyBlue}{E3}{}
      \networkLayer{1.5}{0.4}{2.4}{color=ForestGreen}{}{}
      \networkLayer{1.5}{0.4}{2.9}{color=Goldenrod}{}{}
      
      \networkLayer{1.5}{0.8}{3.5}{color=SkyBlue}{E4}{}
      \networkLayer{1.5}{0.8}{4.4}{color=Goldenrod}{}{}

      \networkLayer{1.5}{0.8}{5.4}{color=SkyBlue}{D1}{}
      \networkLayer{1.5}{0.8}{6.3}{color=Goldenrod}{}{}
      
      \networkLayer{2.0}{0.4}{7.3}{color=Salmon}{}{}
      \networkLayer{2.0}{0.4}{7.8}{color=SkyBlue}{D2}{}
      \networkLayer{2.0}{0.4}{8.3}{color=Goldenrod}{}{}
      
      \networkLayer{2.5}{0.2}{8.9}{color=Salmon}{}{}
      \networkLayer{2.5}{0.2}{9.2}{color=SkyBlue}{D3}{}
      \networkLayer{2.5}{0.2}{9.5}{color=Goldenrod}{}{}
      
      \networkLayer{3.0}{0.1}{9.9}{color=Salmon}{}{}
      \networkLayer{3.0}{0.05}{10.1}{color=SkyBlue}{D4}{}

      \networkLayer{3.0}{0.05}{10.7}{color=Violet!50}{O}{}
    \end{tikzpicture}\hspace{2ex}
    \begin{tikzpicture}[baseline = (current bounding box.east)]
    \matrix [draw=black!33] at (current bounding box.east) {
        \node [shape=rectangle, fill=SkyBlue, label=right:SepConv + ReLU] {}; \\
        \node [shape=rectangle, fill=ForestGreen, label=right:MaxPool] {}; \\
        \node [shape=rectangle, fill=Goldenrod, label=right:BatchNorm] {}; \\
        \node [shape=rectangle, fill=Salmon, label=right:UpSample] {}; \\
        \node [shape=rectangle, fill=Violet!50, label=right:SoftMax] {};
        \\
      };
    \end{tikzpicture}
  }
  \caption{The architecture of the networks used consists of a series
    of fully convolutonal encoding (E1--E4) and decoding (D1--D4) steps. 
    A pixelwise softmax output layer provides the final
    classifications. Network variations differ in the actual number of
    encoding and decoding steps used, filter size and initial depth,
    filter depth multiplication factor and convolution
    stride.}\label{fig:cnn}
\end{figure}

As mentioned before, our approach is based on fully convolutional
semantic segmentation networks. The main structure of our networks is
similar to popular encoder-decoder networks, such as
U-Net~\cite{ronneberger2015u} and
SegNet~\cite{badrinarayanan2017segnet}, mainly following the latter. In
such networks, a first series of convolution layers encode the input
into successively lower resolution but higher dimensional feature
maps, after which a second series of layers decode these maps into a
full-resolution pixelwise classification. This architecture is shown
in Fig.~\ref{fig:cnn}.

SegNet and U-Net both have the property that some information from the
encoder layers are fed into the respective decoder layers of the same
size, either in terms of maxpooling indices, or full feature maps. This
helps overcoming the loss of fine detail caused by the resolution
reduction along the way. As good performance is still possible without these
connections, we do not use those here. They in fact introduce a
significant increase in computation load on our hardware, due to
having to combine tensors in possibly significantly different memory
locations.

Another modification is to use depthwise separable convolution, as
introduced by MobileNets~\cite{howard2017mobilenets}, as a much more
efficient alternative to full 3D~convolution. This is one of the major
contributions to efficiency of our networks, without significantly
decreasing their performance.

To study the trade-off between network minimalism, efficiency and
performance, we create, train and evaluate a number of varieties of
the above network, with different combinations of the following
parameter values:
\begin{enumerate}
\item {\bf Number of Layers (L)} --- $L \in \{3, 4\}$, the number of encoding and
  decoding layers. We always use the same number of encoding and
  decoding layers.
\item {\bf Number of Filters (F)} --- $F \in \{3, 4, 5\}$, the number of filters used in
  the first encoding layer.
\item {\bf Filter Multiplier (M)} --- $M \in \{1.25, 1.5, 2\}$, the factor by which the number
  of filters increases for each subsequent encoding layer, and
  decreases for each subsequent decoding layer.
\item {\bf Convolution Stride (S)} --- $S \in \{1, 2\}$, the stride used in each
  convolution layer.
\end{enumerate}
Larger and smaller values for these parameters have been tried out,
but we only report those here that resulted in networks that were able
to learn the provided task to some degree, but were light enough to be
run on the minimal test hardware. Specific instantiations will be
denoted with $L_xF_yM_zS_w$ with parameter values filled into the
place holders. For instance, $L_3F_4M_{1.25}S_2$ is the network with 3
encoding and decoding layers, \num{4}~features in the first feature map, a
multiplier of 1.25 (resulting in \numlist{4;5;6}~features in each
subsequent layer) and a stride of 1. Not all combinations are valid:
a combination of $L=4$ and $S=2$ would result in invalid feature map
sizes given our input of \num{640x480}~images. The total number of
network types then is 27. Finally, all convolution layers use \num{3x3}~filters, 
padding to have output size the same as input size and no
bias.

\section{Experiments}
\label{sec:experiments}

\begin{figure}[t]
  \centering
  \scriptsize
  \begin{tabular}{@{}c@{}c@{}c@{}c@{}c@{}}
    \includegraphics[width=.2\textwidth]{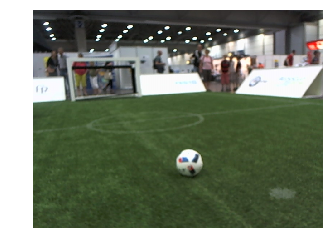} &%
    \includegraphics[width=.2\textwidth]{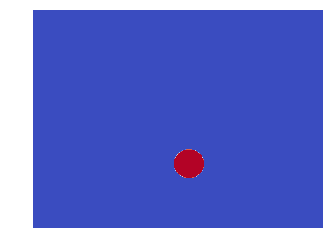} &%
    \includegraphics[width=.2\textwidth]{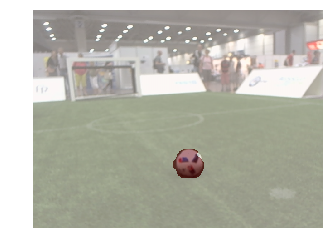} &%
    \includegraphics[width=.2\textwidth]{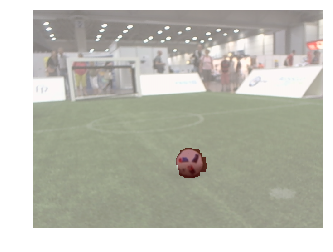} &%
    \includegraphics[width=.2\textwidth]{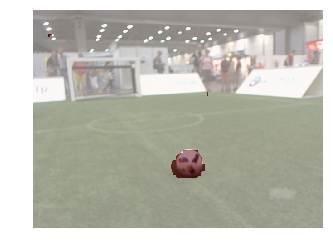} \\[-5pt]%
    \includegraphics[width=.2\textwidth]{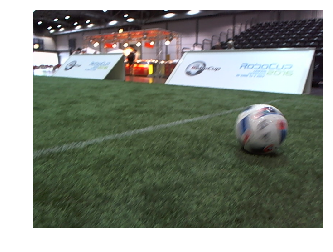} &%
    \includegraphics[width=.2\textwidth]{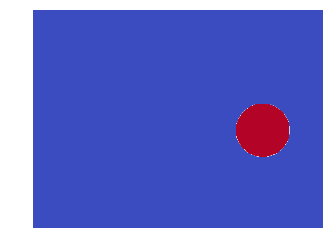} &%
    \includegraphics[width=.2\textwidth]{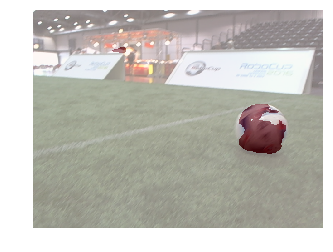} &%
    \includegraphics[width=.2\textwidth]{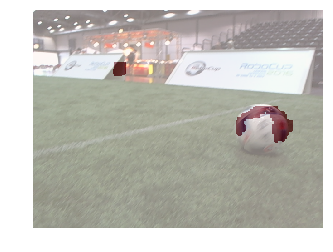} &%
    \includegraphics[width=.2\textwidth]{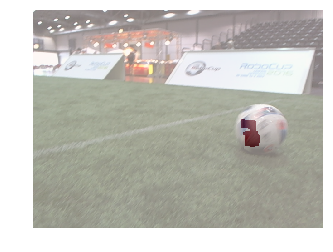} \\[-5pt]
    \includegraphics[width=.2\textwidth]{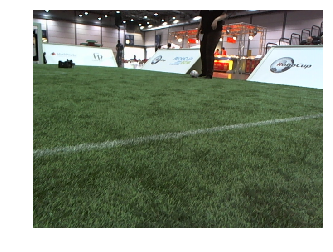} &%
    \includegraphics[width=.2\textwidth]{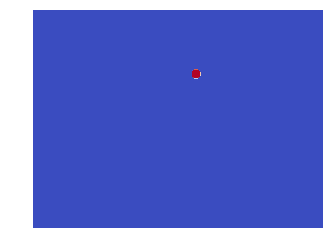} &%
    \includegraphics[width=.2\textwidth]{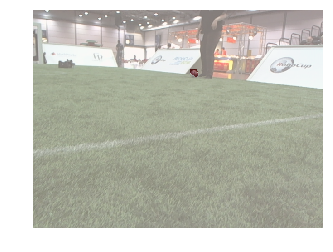} &%
    \includegraphics[width=.2\textwidth]{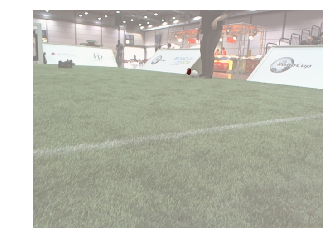} &%
    \includegraphics[width=.2\textwidth]{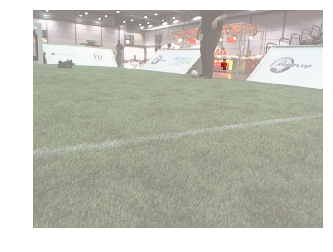} \\

    RGB & Target & $L_4F_5M_2S_1$ & $L_3F_5M_2S_2$ & $L_3F_4M_{1.5}S_2$
  \end{tabular}
  
  \caption{Examples of input, target and segmentation outputs. The
    outputs are respectively of the best stride~1 network, the best
    stride~2 network, and the second best stride~2 network that
    achieves 20 frames per second on QVGA images.}
  \label{fig:exampleimages}
\end{figure}

The networks described in the previous section are trained to segment
ball pixels in images taken by a Kid-Size RoboCup robot on a
competition pitch. Specifically, the image set {\tt bitbots-set00-04}
from the Bit-Bots'
Imagetagger\footnote{\url{https://imagetagger.bit-bots.de/images/imageset/12/}}
was used. It contains 1000 images\footnote{Images taken at the 2016 world
championship in Leipzig, Germany.} with 1003 bounding box ball
annotations. For deriving the target pixel label masks for training the
networks, the rectangular annotations are converted to filled
ellipsoids. Figure~\ref{fig:exampleimages} shows an example of the input images and targets.

We use the TensorFlow library to construct, train and run the
networks. The networks are trained on an NVIDIA GeForce GTX 1080-ti
GPU, with a categorical cross-entropy loss function using stochastic
gradient decent with a starting learning rate of 0.1, a decay factor
of 0.004 and a momentum of 0.9. The dataset is split in a training set
of \num{750}~images, a validation set of 150 and a test set of 100
images. The sets are augmented to double their size by including all
horizontally mirrored images. During training, a batch size of 
\num{10}~images is used. Networks are trained for \num{25}~epochs.

For testing the performance of the networks we map the class
probabilities from the softmax output to discrete class labels and use
this to calculate the commonly used \emph{Intersection over
  Union}~(IoU) score as $\frac{TP}{TP+FP+FN}$, where $TP$ is the
number of true positive ball pixels, $FP$ the number of false
positives and $FN$ the number of false negatives. Due to the extreme
class imbalance, the networks hardly ever predict the probability of a
pixel being part of a ball, $P(B)$, to be above 0.5. This means that
if we use the \emph{most probable} class as final output, the IoU
score often is 0, even though the networks do learn to assign
relatively higher probability at the right pixels. Instead we find the
threshold $\theta^*$ for $P(B)$ that results in the best IoU score for
each trained network.

Finally, since the original goal is to develop networks that can run
on minimal hardware, the networks are run and timed on such hardware,
belonging to a Kid-Size humanoid football robot, namely an
Odroid-XU4. This device is based on a Samsung Exynos~5422 Cortex-A15
with \SI{2}{\giga\hertz} and a Cortex-A7 Octa core CPU, which is the
same as used in some 2015 model flagship smartphones. Before running
the networks, they are optimised using TensorFlow's \emph{Graph
  Transform} tool, which is able to merge several operations, such as
batch normalisation, into more efficient ones. The test program and
TensorFlow library are compiled with all standard optimisation flags
for the ARM Neon platform. We time the networks both on full
\num{640x480}~images and on \num{320x256}~images.

\section{Results}
\label{sec:results}
We firstly evaluate the performance of semantic segmentation networks
trained for the official RoboCup ball. We compare the performance and
runtime of the different network instantiations with each other, as
well as to a baseline segmentation method. This method is based on a
popular fast \emph{lookup table}~(LUT) method, where the table
directly maps pixel values to object classes. To create the table, we
train a \emph{Support Vector Machine}~(SVM) on the same set as the
CNNs to classify pixels. More complex and performant methods may be
chosen, perhaps specifically for the robot football scenario, however
we selected this method to reflect the same workflow of training a
single model on simple pixel data, without injecting domain specific
knowledge. We did improve performance by using input in HSV colour
space and applying grid search to optimise its hyper parameters.

Secondly, we extend the binary case and train the networks for balls
and goal posts, and compare the network performance with the binary
segmentation.

\subsection{Binary segmentation}
\label{sec:results:bin_seg}

\begin{figure}[htb]
  \centering
  \includegraphics[width=.4\textwidth]{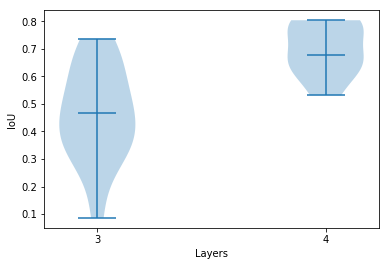}
  \includegraphics[width=.4\textwidth]{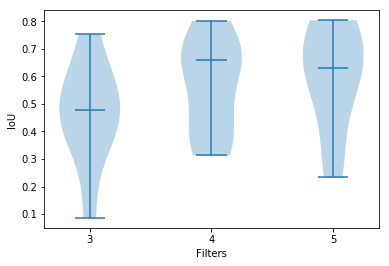}

  \includegraphics[width=.4\textwidth]{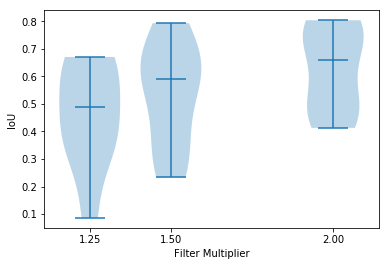}
  \includegraphics[width=.4\textwidth]{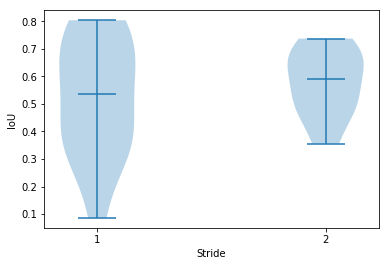}

  \caption{Performance distributions for each network
    parameter. Horizontal bars show the minimum, median and maximum
    scores. The lighter area indicates the distribution over the
    scores through a kernel-density estimation.}
  \label{fig:perfdists}
\end{figure}

We first analyse the segmentation quality of the set of networks
and the influence of their parameters on their performance. The best
network is $L_4F_5M_2S_1$ with a best IoU of 0.804. As may be
expected, this is the one with the most layers, most filters, highest
multiplication factor and densest stride. The least performant
network is one of the simplest in terms of layers and features:
$L_3F_3M_{1.25}S_1$ with a best IoU of 0.085. Perhaps surprisingly the
version of that network with stride~2 manages to perform better, with
a score of 0.39. Figure~\ref{fig:perfdists} shows the distributions of
performance over networks with the same values for each parameter. One
can see that overall more complex networks score higher, but that the
median network with stride~2 performs better than the median with
stride~1.

\begin{figure}
  \centering
  \includegraphics[width=.45\textwidth]{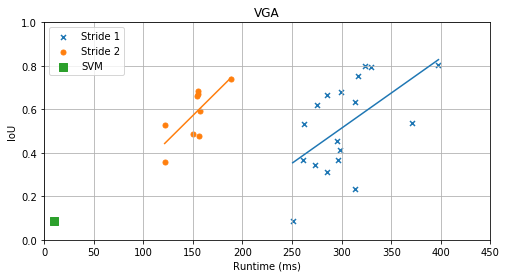} \includegraphics[width=.45\textwidth]{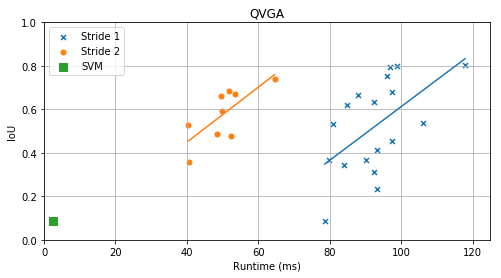}

  \caption{IoU against runtime per image. The mean of the runtime is
    taken over 100 iterations. Results for networks with stride~1 and
    stride~2 are split and plotted with blue crosses and orange
    circles respectively. Additionally, for each group the linear fit
    to the data is shown. The baseline score of the SVM is marked as a
    green square. Left: Full VGA (\num{640x480}), right: QVGA
    (\num{320x256}); note the different timescales.}
  \label{fig:runtime}
\end{figure}

Figure~\ref{fig:runtime} compares the runtime and IoU scores for all
networks. The data points are grouped by stride, resulting in clearly
distinct clusters: as expected the networks with stride~2 have a
significantly lower runtime requirement. The timings run from
\SIrange{121}{397}{\milli\second} per full \num{640x480} resolution
frame, which is equivalent to approximately \SIlist{8; 2.5}{\fpstext},
respectively.

The best performing network in terms of IoU is also the least efficient one,
but the second best is a full \SI{74}{\milli\second} faster with a drop in IoU of only
0.003. The linear fits show that there is indeed a trend within each
cluster of better performance given the runtime, but it is clear that
this is not generally the case: networks with similar runtimes can
vary greatly in achieved performance.

The SVM-based LUT method, though being significantly faster, scores
well below most networks, with an IoU of 0.085. This is because such a
pixel-by-pixel method does not consider the surrounding area and thus
has no way to discern pixels with similar colours, resulting in many
false positives for pixels that have colours that are also found in
the ball. In contrast, the CNNs can perform much more informed
classification by utilising the receptive field around each pixel.

\setlength{\tabcolsep}{8pt}
\begin{table}[htb]
  \scriptsize
\centering
\caption{Segmentation scores and runtimes obtained by networks}
\label{tab:fulldata}
\begin{tabular}{@{}lllllrrr@{}} \toprule
  Layers & Filters & Mult & Stride & $\theta^*$ & IoU & Time (ms)    &  \\
         &         &      &        &     &            & \num{640x480}      & \num{320x256} \\ \midrule
  3         & 3          & 1.25        & 1      & 0.07     & 0.086     & 250 & 78     \\
  3         & 3          & 1.25        & 2      & 0.07     & 0.356     & 121 & 40     \\
  3         & 3          & 1.5        & 1      & 0.30     & 0.366     & 261 & 79     \\
  3         & 3          & 1.5        & 2      & 0.25     & 0.529     & 121 & 40     \\
  3         & 3          & 2        & 1      & 0.33     & 0.456     & 295 & 97     \\
  3         & 3          & 2        & 2      & 0.31     & 0.478     & 156 & 52     \\
  3         & 4          & 1.25        & 1      & 0.23     & 0.343     & 273 & 84     \\
  3         & 4          & 1.25        & 2      & 0.31     & 0.487     & 149 & 48     \\
  3         & 4          & 1.5        & 1      & 0.16     & 0.313     & 285 & 92     \\
  3         & 4          & 1.5        & 2      & 0.28     & 0.682     & 155 & 51     \\
  3         & 4          & 2        & 1      & 0.31     & 0.413     & 297 & 93     \\
  3         & 4          & 2        & 2      & 0.26     & 0.660     & 154 & 49     \\
  3         & 5          & 1.25        & 1      & 0.22     & 0.365     & 296 & 90     \\
  3         & 5          & 1.25        & 2      & 0.25     & 0.671     & 155 & 53     \\
  3         & 5          & 1.5        & 1      & 0.49     & 0.233     & 312 & 93     \\
  3         & 5          & 1.5        & 2      & 0.25     & 0.590     & 156 & 49     \\
  3         & 5          & 2        & 1      & 0.42     & 0.538     & 370 & 106    \\
  3         & 5          & 2        & 2      & 0.24     & \bf 0.738     & 188 & 64     \\ \midrule
  4         & 3          & 1.25        & 1      & 0.43     & 0.531     & 262 & 80     \\
  4         & 3          & 1.5        & 1      & 0.39     & 0.620     & 274 & 84     \\
  4         & 3          & 2        & 1      & 0.49     & 0.754     & 316 & 96     \\
  4         & 4          & 1.25        & 1      & 0.50     & 0.666     & 285 & 87     \\
  4         & 4          & 1.5        & 1      & 0.23     & 0.678     & 299 & 97     \\
  4         & 4          & 2        & 1      & 0.36     & 0.801     & 323 & 98     \\
  4         & 5          & 1.25        & 1      & 0.74     & 0.632     & 313 & 92     \\
  4         & 5          & 1.5        & 1      & 0.41     & 0.794     & 329 & 96     \\
  4         & 5          & 2        & 1      & 0.46     & \bf 0.804     & 397 & 117    \\ \midrule
  SVM & & & & & 0.085 & 10 & 3 \\ \bottomrule
\end{tabular}
\end{table}

From this figure we can conclude the same as from
Fig.~\ref{fig:perfdists}, that introducing a stride of 2 does not
significantly reduce performance, but with the addition that it does
make the network run significantly faster. The best network with
stride~2, $L_3F_5M_2S_2$, has an IoU of only 0.066 less than the best
network (a drop of just 8\%), but runs over twice as fast, at more
than 5 frames per second.  On the lower resolution of \num{320x256}
the best stride~2 networks achieve frame rates of
\SIrange{15}{20}{\fpstext}. Table~\ref{tab:fulldata} lists the results
for all networks.

\subsection{Multi-class segmentation}

Binary classification is too limited for robotic football, or other
real world scenarios. To study the more general usability of our
method, we extend the binary-class segmentation case from
Sect.~\ref{sec:results:bin_seg}. The same dataset as before is used,
but with additionally goalposts annotated as a third class. We
selected the best stride~1 and best stride~2 networks to train. These
two networks are kept the same, except for an additional channel added
to the last decoding layer and to the softmax layer.

We found it to be difficult for the networks to successfully learn to
segment the full goalpost, not being able to discern higher sections
above the field edge from parts of the background. Combined with the
fact that the robots typically use points on the ground, as they are
able to better judge their distance, we select the bottom of the
goalposts by labelling a circle with a radius of \num{20}~pixels in the
target output where the goalposts touch the field in the images.

\begin{figure}[t]
  \centering
  \scriptsize
  \begin{tabular}{@{}c@{}c@{}c@{}c@{}}
    \includegraphics[width=.2\textwidth]{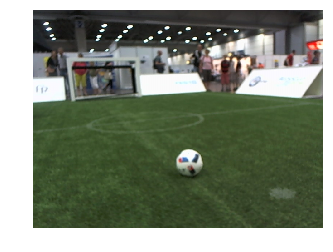} &%
    \includegraphics[width=.2\textwidth]{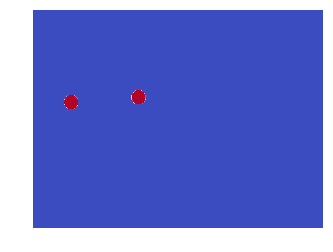} &%
    \includegraphics[width=.2\textwidth]{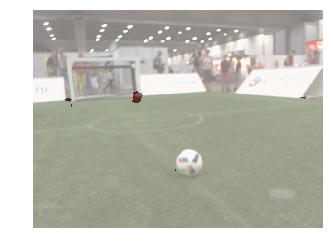} &%
    \includegraphics[width=.2\textwidth]{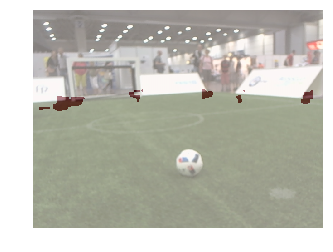}\\[-5pt]%
    \includegraphics[width=.2\textwidth]{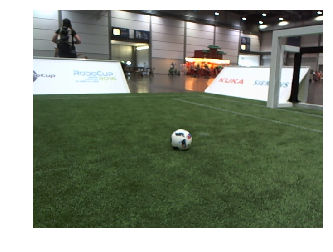} &%
    \includegraphics[width=.2\textwidth]{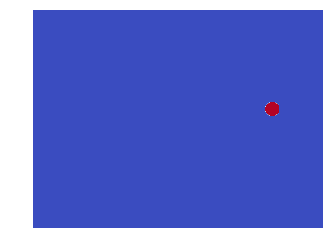} &%
    \includegraphics[width=.2\textwidth]{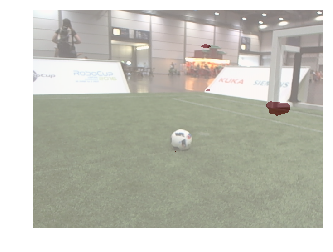} &%
    \includegraphics[width=.2\textwidth]{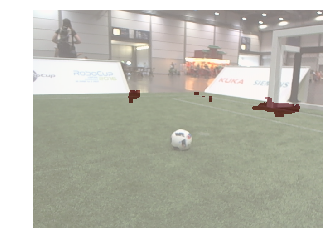}\\[-5pt]%
    \includegraphics[width=.2\textwidth]{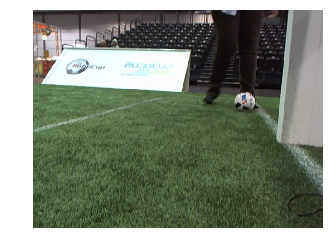} &%
    \includegraphics[width=.2\textwidth]{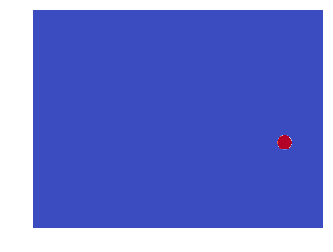} &%
    \includegraphics[width=.2\textwidth]{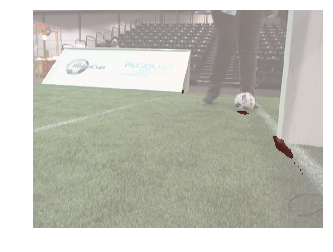} &%
    \includegraphics[width=.2\textwidth]{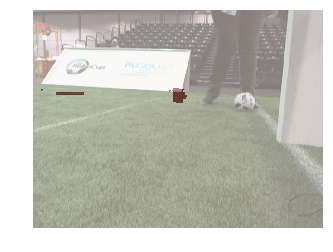}\\%

    RGB & Target & $L_4F_5M_2S_1$ & $L_3F_5M_2S_2$
  \end{tabular}
  
  \caption{Examples of input, target and segmentation outputs for goalposts for the two trained networks.}
  \label{fig:exampleimages_posts}
\end{figure}

Because of the additional difficulty of learning features for an extra
class, the IoU score for the ball class dropped slightly for both networks:
to 0.754 for $L_4F_5M_2S_1$ and to 0.708 for $L_3F_5M_2S_2$, compared
to 0.804 and 0.738 in the binary task. The scores reached for the
goalpost class were 0.273 and 0.102 respectively. Although this does
not reach the same level as for the ball class, the networks are still able
to mark out the bottom of the goal posts in the examples shown in
Fig.~\ref{fig:exampleimages_posts}. Because only some additional
operations in the last layers, the run times are comparable to the
binary equivalents: 414 and 196 milliseconds.

The worse scores on the goalposts are mostly due to false positives,
either marking too large an area or mislabelling other objects,
especially in the case of the stride~2 network. Several reasons
contribute to this. Firstly, the data used is typical for an image
sequence of a robot playing a game of football, where it most of the
time is focused on the ball. This results in goals being less often
visible, and thus in the data being unbalanced: at least one ball is
visible in all \num{1000}~images, whereas at least one goal post is
visible in only \num{408}~images. Secondly, goal posts are less
feature-rich, so more difficult to discern from other
objects. Finally, our annotation method does not mark a
well-distinguished area, making it harder for the networks to predict
its exact shape. Further research is required to alleviate these
issues and improve performance, however the results obtained here
provide evidence that our approach can handle multi-class segmentation
with only little performance reduction.

\section{Conclusions}
\label{sec:conclusions}
We have developed a minimal deep learning semantic segmentation
architecture to be run on minimal hardware. We have shown that such a
network can achieve good performance in segmenting the ball, and show
promising results for additionally detecting goalposts, in a RoboCup
environment with a useful frame rate. Table~\ref{tab:compare} lists the
resolutions and frame rates reported by other authors alongside
ours. It must be noted that a direct comparison is difficult, because
of the different outputs of the used CNNs and the different robot
platforms, but our approach is the only one that has all of the
following properties:
\begin{itemize}
\item Processes full VGA images at \SI{5}{\fps} and QVGA at
  \SIrange{15}{20}{\fps}\footnote{Actually a resolution slightly
    larger than QVGA is used, as the stride~2 networks require
    dimensions that are multiples of 32.}
\item Can handle multiple image dimensions without retraining
\item Does not require task specific knowledge to achieve high frame rate
\item Achieves all this on minimal mobile hardware
\end{itemize}

For achieving full object localisation as given by the other solutions,
additional steps are still required. However, because the output of
the semantic segmentation is in the same form as a lookup table based
labelling approach, any already existing methods built on top of such
a method can directly be reused. For instance, an efficient---and
still task agnostic---connected component based method previously
developed by us readily fits onto the architecture outlined here and
performs the final object detection step within only \SIrange{1}{2}{\milli\second}.

\setlength{\tabcolsep}{4pt}
\begin{table}[]
  \scriptsize
\centering
\caption{Reported resolutions and frame rate of DL application in RoboCup domain}
\label{tab:compare}
\begin{tabular}{lrrr} \toprule
  & Resolution & FPS & Notes \\ \midrule
  Speck et al.~\cite{speck2016ball} & \num{200x150} & 3 & Predict separate ball x and y \\
  Albani et al.~\cite{albani2016deep} & N/A & 11--22 & Task dependent region proposal \\
  Cruz et al.~\cite{cruz2017using} & \num{24x24} & 440 & Task dependent region proposal \\
  Javadi et al.~\cite{javadi2017humanoid} & N/A & 240 & no loss: \SI{6}{\fps}; task dependent \\
  Da Silva et al.~\cite{da2017towards} & \num{110x110} & 8 & Predict end-to-end desired action \\
  Hess et al.~\cite{hess2017stochastic} & \num{32x32} & 50 & Focus on generation of training data \\
  Schnekenburger et al.~\cite{schnekenburger2017detection} & \num{640x512} & 111 & GTX-760 GPU; \SI{19}{\fps} on i7 CPU \\ \midrule
  Ours & \num{640x480} & 5 & $L_3F_5M_2S_2$ \\
       & \num{320x256} & 15 & $L_3F_5M_2S_2$; $L_3F_4M_{1.5}S_2$: 20 fps \\ \bottomrule
\end{tabular}
\end{table}

By delaying the use of task dependent methods, one actually has an
opportunity to optimise the segmentation output for such methods, by
varying the threshold used to determine the final class pixels. For
specific use cases it may be desirable to choose a threshold that
represents a preference for either high true positive rate (recall),
e.g.\ when a robot's vision system requires complete segmentation of
the ball and/or it has good false-positive filtering algorithms, or
for low false positive rate (fall-out), e.g., when it can work well
with only partly segmented balls, but struggles with too many false
positives.


\begin{thebibliography}{10}
\providecommand{\url}[1]{\texttt{#1}}
\providecommand{\urlprefix}{URL }
\providecommand{\doi}[1]{https://doi.org/#1}

\bibitem{albani2016deep}
Albani, D., Youssef, A., Suriani, V., Nardi, D., Bloisi, D.D.: A deep learning
  approach for object recognition with nao soccer robots. In: RoboCup 2016:
  Robot World Cup XX. pp. 392--403. Springer (2016)

\bibitem{allali2017rhoban}
Allali, J., Deguillaume, L., Fabre, R., Gondry, L., Hofer, L., Ly, O., N'Guyen,
  S., Passault, G., Pirrone, A., Rouxel, Q.: Rhoban football club: Robocup
  humanoid kid-size 2016 champion team paper. In: RoboCup 2017: Robot World Cup
  XXI (2017)

\bibitem{badrinarayanan2017segnet}
Badrinarayanan, V., Kendall, A., Cipolla, R.: Segnet: A deep convolutional
  encoder-decoder architecture for image segmentation. IEEE transactions on
  pattern analysis and machine intelligence  \textbf{39}(12),  2481--2495
  (2017)

\bibitem{cheng1997real}
Cheng, G., Zelinsky, A.: Real-time vision processing for a soccer playing
  mobile robot. In: Robot Soccer World Cup. pp. 144--155. Springer (1997)

\bibitem{cruz2017using}
Cruz, N., Lobos-Tsunekawa, K., Ruiz-del Solar, J.: Using convolutional neural
  networks in robots with limited computational resources: Detecting nao robots
  while playing soccer  (2017)

\bibitem{da2017towards}
Da~Silva, I.J., Vilao, C.O., Costa, A.H., Bianchi, R.A.: Towards robotic
  cognition using deep neural network applied in a goalkeeper robot. In:
  Robotics Symposium (LARS) and 2017 Brazilian Symposium on Robotics (SBR),
  2017 Latin American. pp.~1--6. IEEE (2017)

\bibitem{he2016deep}
He, K., Zhang, X., Ren, S., Sun, J.: Deep residual learning for image
  recognition. In: Proceedings of the IEEE conference on computer vision and
  pattern recognition. pp. 770--778 (2016)

\bibitem{hess2017stochastic}
Hess, T., Mundt, M., Weis, T., Ramesh, V.: Large-scale stochastic scene
  generation and semantic annotation for deep convolutional, neural network
  training in the robocup spl. In: RoboCup 2017: Robot World Cup XXI (2017)

\bibitem{howard2017mobilenets}
Howard, A.G., Zhu, M., Chen, B., Kalenichenko, D., Wang, W., Weyand, T.,
  Andreetto, M., Adam, H.: Mobilenets: Efficient convolutional neural networks
  for mobile vision applications. arXiv preprint arXiv:1704.04861  (2017)

\bibitem{javadi2017humanoid}
Javadi, M., Azar, S.M., Azami, S., Shiry, S., Ghidary, S.S., Baltes, J.:
  Humanoid robot detection using deep learning: A speed-accuracy tradeoff. In:
  RoboCup 2017: Robot World Cup XXI (2017)

\bibitem{lecun2015deep}
LeCun, Y., Bengio, Y., Hinton, G.: Deep learning. nature  \textbf{521}(7553),
  ~436 (2015)

\bibitem{rastegari2016xnor}
Rastegari, M., Ordonez, V., Redmon, J., Farhadi, A.: Xnor-net: Imagenet
  classification using binary convolutional neural networks. In: European
  Conference on Computer Vision. pp. 525--542. Springer (2016)

\bibitem{ronneberger2015u}
Ronneberger, O., Fischer, P., Brox, T.: U-net: Convolutional networks for
  biomedical image segmentation. In: International Conference on Medical image
  computing and computer-assisted intervention. pp. 234--241. Springer (2015)

\bibitem{schnekenburger2017detection}
Schnekenburger, F., Scharffenberg, M., W{\"u}lker, M., Hochberg, U., Dorer, K.:
  Detection and localization of features on a soccer field with feedforward
  fully convolutional neural networks (fcnn) for the adult-size humanoid robot
  sweaty. In: Proceedings of the 12th Workshop on Humanoid Soccer Robots,
  IEEE-RAS International Conference on Humanoid Robots, Birmingham (2017)

\bibitem{silver2016mastering}
Silver, D., Huang, A., Maddison, C.J., Guez, A., Sifre, L., Van Den~Driessche,
  G., Schrittwieser, J., Antonoglou, I., Panneershelvam, V., Lanctot, M.,
  et~al.: Mastering the game of go with deep neural networks and tree search.
  nature  \textbf{529}(7587),  484--489 (2016)

\bibitem{speck2016ball}
Speck, D., Barros, P., Weber, C., Wermter, S.: Ball localization for robocup
  soccer using convolutional neural networks. In: RoboCup 2016: Robot World Cup
  XX. pp. 19--30. Springer (2016)

\bibitem{zhou2016dorefa}
Zhou, S., Wu, Y., Ni, Z., Zhou, X., Wen, H., Zou, Y.: Dorefa-net: Training low
  bitwidth convolutional neural networks with low bitwidth gradients. arXiv
  preprint arXiv:1606.06160  (2016)

\end{thebibliography}
\end{document}